\documentclass[]{article} \usepackage[T1]{fontenc}
\usepackage{graphicx}
\usepackage{tikz}
\usetikzlibrary{arrows,positioning,shapes.misc,shapes.geometric}
\usepackage{todonotes}
\usepackage{xspace}
\usepackage{comment}
\usepackage{amsmath}
\usepackage{enumitem}
\usepackage{apxproof}

\tikzset{
>=stealth',
args/.style={circle,draw=black}
}

\newtheorem{definition}{Definition}
\newtheorem{example}{Example}
\newtheorem{proposition}{Proposition}
\newtheorem{corollary}{Corollary}

\newtheoremrep{propositionAp}[proposition]{Proposition}
\newtheoremrep{factAp}{Fact}
\newtheoremrep{lemmaAp}{Lemma}
\newtheoremrep{corollaryAp}[corollary]{Corollary}

					\newcommand{\true}{\ensuremath{1}}						\newcommand{\false}{\ensuremath{0}}						\newcommand{\undec}{\ensuremath{\mathsf{U}}}										\newcommand{\atoms}{\ensuremath{\mathsf{At}}}

\newcommand{\Vc}{{\cal V}\xspace}

\newcommand{\BN}{\textsf{BN}\xspace}
\newcommand{\BNs}{\textsf{BNs}\xspace}

\newcommand{\ADF}{\textsf{ADF}\xspace}
\newcommand{\ADFs}{\textsf{ADFs}\xspace}

\newcommand{\preferred}{\textsf{prf}}
\newcommand{\grounded}{\textsf{grnd}}
\newcommand{\admissible}{\textsf{adm}}
\newcommand{\complete}{\textsf{cmp}}
\newcommand{\stable}{\textsf{stb}}
\newcommand{\twoV}{\textsf{2V}}

\newcommand{\modelSettwo}[1]{{\ensuremath{\mathsf{Mod}^2(#1)}}}
\newcommand{\possibleWorld}{\omega}
\title{Abstract Dialectical Frameworks are Boolean Networks (full version)}
\author{Jesse Heyninck$^{1,2}$, 
Matthias Knorr$^{3}$ and
Jo\~ao Leite$^{3}$\\
{\small $^{1}$Open Universiteit, the Netherlands}\\
{\small$^{2}$ University of Cape Town, South-Africa}\\
{\small $^{3}$ NOVA LINCS, NOVA University Lisbon }
}

\begin{document}

\maketitle              \begin{abstract}

Abstract dialectical frameworks are a unifying model of formal argumentation, where argumentative relations between arguments are represented by assigning acceptance conditions to atomic arguments. 
Their generality allow them to cover a number of different approaches with varying forms of representing the argumentation structure. 
Boolean regulatory networks are used to model the dynamics of complex biological processes, taking into account the interactions of biological compounds, such as proteins or genes. 
These models have proven highly useful for comprehending such biological processes, allowing to reproduce known behaviour and testing new hypotheses and predictions \textit{in silico}, for example in the context of new medical treatments.
While both these approaches stem from entirely different communities, it turns out that there are striking similarities in their appearence.
In this paper, we study the relation between these two formalisms revealing their communalities as well as their differences, and introducing a correspondence that allows to establish novel results for the individual formalisms.
\end{abstract}

\setcounter{footnote}{0} 

\section{Introduction}\label{sec:intro}

\begin{figure}[!t]
	\centering
	\ \ \ \ \ \	\includegraphics[scale=0.26]{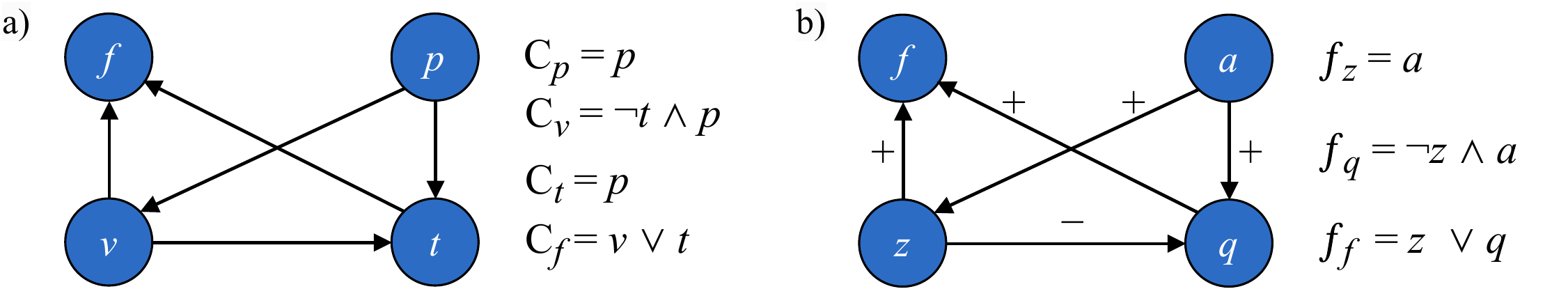}
	\caption{\label{figure:intro} Examples of a) an \ADF for Example \ref{example:travelling}, and b) a \BN for Example \ref{example:mussels}.}
\end{figure}

Formal argumentation is one of the major approaches to knowledge representation and reasoning. In the seminal paper \cite{Dung1995}, Dung introduced \emph{abstract argumentation frameworks}, conceived as directed graphs where nodes represent arguments and edges between nodes represent attacks. Their meaning is given by so-called \emph{argumentation semantics} that determine which sets of arguments can be reasonably upheld together given such an argumentation graph.
Various authors have since remarked that other relations between arguments are worth consideration, such as, for example, a dual \emph{support} relation as in \emph{bipolar argumentation frameworks} \cite{cayrol2005acceptability}. 
The last decades witnessed a proliferation of extensions of the original formalism that has often made it hard to compare the dialects of the different resulting argumentation formalisms. 
In an attempt to cope with the resulting number of dialects and unify them, \emph{abstract dialectical frameworks} (in short, \ADFs) were introduced \cite{brewka2013abstract}.
Just like abstract argumentation frameworks, 
\ADFs are also directed graphs. However, in \ADFs, 
edges between nodes do not necessarily represent attacks, but can encode \emph{any} relationship between arguments. Such generality is achieved by associating an \emph{acceptance condition} with each argument, represented as a Boolean formula over the parents of the argument, expressing the conditions under which the argument can be accepted. \ADFs offer a general framework for argumentation-based inference as they are able to capture all of the major semantics of abstract argumentation \cite{polberg2016understanding}, and even normal logic programs \cite{brewka2013abstract}.
The following example illustrates a simple \ADF.
\begin{example}\label{example:travelling}
You are considering your travel plans for the upcoming conference summer. If you manage to write a paper, it will be suitable for a conference in $t$exas or in $v$ietnam. If you submit the paper to the conference in Vietnam, you cannot submit it to the conference in Texas (but the conference in Texas does allow for submission of paper submitted elsewhere). Both conferences will require you to apply for travelling $f$unds. This example can be expressed formally in the \ADF given in Fig.~\ref{figure:intro}.a). Interactions between atoms are expressed by arrows (e.g.\ writing a paper influences attendance of a conference in both Vietnam and Texas), and the acceptance conditions express these interactions more precisely. E.g., $C_v=\lnot t\land p$ expresses conference attendance in Vietnam is possible if one has a paper and did not submit it to the conference in Texas.
\end{example}

In systems biology, \emph{Biological regulatory networks} encode interactions between biological specimens or compounds, such as proteins or genes, and their interactions, to acquire a better understanding of the complex processes that take place in cells, as doing so may lead to discoveries and new theories about living organisms. 
To abstract from actual concentration values, and use thresholds to represent whether a compound is active or inactive, Logical models \cite{kaufflogic,THOMAS1973563} are often used instead of quantitative models. 
Because of that, they require far less information than quantitative models and are therefore more adequate to deal with incomplete, imprecise, and noisy information regarding the biological system. Among these, Boolean (logical) models or networks (\BNs), have been extensively used to reproduce known behaviour and test new hypotheses \textit{in silico}, e.g., as models of gene regulation networks and other biological systems \cite{SalinasGA22}. 
\begin{example}\label{example:mussels}
Consider the following toy scenario of a biosphere where four species are potentially present: native $q$uagga mussels, invasive $z$ebra mussels,  $a$lgae, and $f$ish. These species interact as follows:
fish feed on mussels,
zebra-mussels outcompete quagga,
and both kinds of mussels feed on algae.
These interactions can be represented using the biological network in Fig.~\ref{figure:intro}.b). Notice that this is a simplified representation of a biological network, meant merely to ease understanding. Influences are represented by arrows, marked with a ``$+$'' if the influence is positive (e.g. since fish feed on mussels), and a ``$-$'' if the influence is negative (e.g.\ since zebra mussels outcompete quagga mussels). Furthermore, the precise nature of these interactions is encoded by Boolean functions on the right-hand side of the figure. For example, $f_q=\lnot z\land a$ expresses that quagga mussels will be alive if there are no zebra-mussels and there are algae.
\end{example}

These examples show striking similarities between these models of two very different subject matters: \ADFs and \BNs. Syntactically, they both use directed graphs to encode interactions, and boolean formulae to express the precise nature of such interactions. The open question is whether such similarities extend from the syntactic to the semantic level. If that were the case, it would open up a fruitful ground for the cross-fertilization of both fields, both theoretically, where established results from one could provide new insights for the other, but also practically, e.g., by allowing implementations for one to be used by the other.

In this paper, we compare \ADFs and \BNs. After formally introducing \ADFs in Sec.~\ref{sec:adf}, in Sec.~\ref{sec:bn} we introduce \BNs and show that the similarities with \ADFs extend from the syntactic to the semantic level, by pointing in a formally precise way what are the equivalent notions, but also what the differences are. Our results show that \ADFs are essentially \BNs, creating a fruitful ground for investigating synergies, as illustrated in Sec.~\ref{section:results:from:bns}. We then conclude in Sec.~\ref{sec:conclusion}.

\section{Abstract Dialectical Frameworks}\label{sec:adf}

We recall \ADFs following loosely the notation from \cite{brewka2013abstract}. 
An \ADF $D$ is a tuple $D=(\atoms,L,C)$ where $\atoms$ is a finite set of atoms, representing arguments or statements, $L\subseteq \atoms\times \atoms$ is a set of links, representing dependencies or attacks from one argument against another, and $C=\{C_{s}\}_{s\in \atoms}$ is a set of total functions (also called acceptance functions) $C_{s}:2^{par_{D}(s)}\rightarrow \{\true,\false\}$ for each $s\in \atoms$ with $par_{D}(s)=\{s'\in \atoms\mid (s',s)\in L\}$ and truth values true ($\true$) and false ($\false$). 
An acceptance function $C_{s}$ defines the cases when the statement $s$ can be accepted (is true),
depending on the acceptance status of its parents in $D$. 
We often identify an acceptance function $C_{s}$ by its equivalent \emph{acceptance condition} which models the acceptable cases as a propositional formula over $\atoms$ and the usual Boolean connectives $\wedge$ (\emph{and}), $\vee$ (\emph{or}), $\neg$ (\emph{negation}) and $\rightarrow$ (\emph{material implication}). Also, the set of links of $D$ is completely determined by $C_{s}$ and sometimes left implicit.

\begin{example}\label{exa:ADF:1}\label{ex:ADF1}
In Ex.~\ref{example:travelling}, we find an \ADF $D=(\{f,p,t,v\},L,C)$ with $L$ and $C$ as in Fig~\ref{figure:intro}.a). Here, we omit reflexive arrows (e.g.\ $(p,p)$) to avoid clutter. An acceptance condition like $C_v=\lnot t\land p$ can be read as ``$v$ is accepted if $t$ is not accepted and $p$ is accepted.
\end{example}
Interpretations can be used to formally assign meaning to these acceptance conditions.
An \emph{interpretation} (also called \emph{possible world}) $\possibleWorld$ is a function
$\possibleWorld:\atoms\rightarrow\{\true,\false\}$. Let $\Vc^2(\atoms)$ denote the set of all interpretations for \atoms.
We simply write $\Vc^2$ if the set of atoms is implicitly given.
An interpretation $\possibleWorld$ \emph{satisfies} (or is a \emph{model} of) an atom $a\in\atoms$, denoted by $\possibleWorld\models a$, if and only if $\possibleWorld(a)=\true$. The satisfaction relation $\models$ is extended to formulas as usual.
Then, an interpretation $\possibleWorld$ is a \emph{two-valued model} of an \ADF $D$ if, for all $s\in \atoms$, $\possibleWorld\models s$ iff $\possibleWorld\models C_s$.
For sets of formulas $\Phi$, we also define $\possibleWorld\models \Phi$ if and only if $\possibleWorld\models\phi$ for every $\phi\in\Phi$, and the set of models $\modelSettwo{\Phi}=\{\omega\in {\cal V}^2(\atoms)\mid \possibleWorld\models \Phi\}$ for every set of formulas $\Phi$.
A set of formulas $\Phi_{1}$ \emph{entails} another set of formulas $\Phi_{2}$, denoted by $\Phi_{1}\vdash \Phi_{2}$, if $\modelSettwo{\Phi_{1}}\subseteq\modelSettwo{\Phi_{2}}$.
A formula $\phi$ is a tautology if $\modelSettwo{\phi}={\cal V}^2(\atoms)$ and inconsistent if $\modelSettwo{\phi}=\emptyset$.
We compare two possible worlds $\omega_1$ and $\omega_2$ by: $\omega_1\leq \omega_2$ iff for every $\alpha\in \atoms$, $\omega_1(\alpha)=\true$ implies $\omega_2(\alpha)=\true$.

Commonly though, an \ADF $D=(\atoms,L,C)$ is interpreted through 3-valued interpretations $\nu:\atoms\rightarrow \{\true,\false,\undec\}$ adding truth value undecided ($\undec$). We denote the set of all 3-valued interpretations over $\atoms$ by $\Vc^3(\atoms)$.
We define the information order $<_i$ over $\{\true,\false,\undec\}$ by making $\undec$ the minimal element: $\undec<_i \true$ and $\undec<_i\false$, and $\dagger\leq_i \ddagger$ iff $\dagger <_i \ddagger$ or $\dagger=\ddagger$ for any $\dagger,\ddagger\in \{\true,\false,\undec\}$. This order is lifted point-wise as follows (given $\nu,\nu'\in\Vc^3(\atoms)$): $\nu\leq_i \nu'$ iff $\nu(s)\leq_i \nu'(s)$ for every $s\in \atoms$. 
The set of two-valued interpretations extending a 3-valued interpretation $\nu$ is defined as $[\nu]^2=\{\omega\in \Vc^2(\atoms)\mid \nu\leq_i \omega\}$. 
Given a set of 3-valued interpretations $V\subseteq \Vc^3(\atoms)$, $\sqcap_i V$ is the 3-valued interpretation defined via $\sqcap_i V(s)=\dagger$ if for every $\nu\in V$, $\nu(s)=\dagger$, for any $\dagger\in \{\true,\false,\undec\}$, and $\sqcap_i V(s)=\undec$ otherwise.

All major semantics of \ADFs single out three-valued interpretations in which the truth value of every atom $s\in \atoms$ is, in some sense, in alignment or agreement with the truth value of the corresponding condition $C_s$. 
The $\Gamma$-function enforces this intuition by mapping an interpretation $\nu$ to a new interpretation $\Gamma_D(\nu)$, which assigns to every atom $s$ exactly the truth value assigned by $\nu$ to $C_s$, i.e.: 
\[
\Gamma_D(\nu): \atoms\rightarrow \{\true,\false,\undec\} \text{ where }s\rightarrow \sqcap_i \{\omega(C_s)\mid\omega\in [\nu]^2\}.
\]
We also need to define the reduct $D^\nu$ of $D$ given $\nu$, i.e., $D^\nu=(\atoms^\nu,L^\nu,C^\nu)$ with: (i) $\atoms^\nu=\{s\in\atoms\mid \nu(\atoms)=\true\}$, (ii) $L^\nu=L\cap (\atoms^\nu\times \atoms^\nu)$, and (ii) $C^\nu= \{C_s[\{\phi\mid \nu(\phi)=\false\}/\false]\mid s\in \atoms^\nu\}$, where $C_s[\phi/\psi]$ is obtained by substituting every occurrence of $\phi$ in $C_s$ by $\psi$.
\begin{definition}
	Let $D$ be an \ADF with $\nu\in\Vc(\atoms)$ a 3-valued interpretation. Then, 
$\nu$ is \emph{admissible for $D$} iff $\nu\leq_i\Gamma_D(\nu)$;
$\nu$ is \emph{complete for $D$} iff $\nu=\Gamma_D(\nu)$; 
$\nu$ is \emph{preferred for $D$} iff $\nu$ is $\leq_i$-maximal among all admissible models;
$\nu$ is \emph{grounded for $D$} iff $\nu$ is $\leq_i$-minimal among all complete models;
and $\nu$ is \emph{stable} iff $\nu$ is a two-valued model of $D$ and $\{s\in \atoms\mid \nu(s)=\true\}=\{s\in \atoms\mid w(s)=\true\}$ where $w$ is the grounded model of $D^{\nu}$.
We denote by $\twoV(D)$, $\admissible(D)$, $\complete(D)$, $\preferred(D)$, $\grounded(D)$, and $\stable(D)$ the sets of two-valued, admissible, complete, preferred, grounded, and stable models of $D$. \end{definition}
It was been shown that $\stable(D)\subseteq \twoV(D) \subseteq \preferred(D) \subseteq \complete(D)\subseteq \admissible(D)$ as well as $\grounded(D)\subseteq
\complete(D)$.

\begin{example}[Ex.\ \ref{example:travelling} ctd.]\label{ex:travelMods}
The \ADF in Ex.\  \ref{example:travelling} has three complete models $\nu_1$, $\nu_2$, $\nu_3$ with: $\nu_1(p)=\true$, $\nu_1(v)=\false$, $\nu_1(t)=\true$ and $\nu_1(f)=\true$;  $\nu_2(s)=\false$ and $\nu_3(s)=\undec$ for all $s\in\atoms$.
	An admissible model that is not complete is $\nu_4$ with $\nu_4(p)=\true$, $\nu_4(v)=\nu_4(t)=\nu_4(f)=\undec$.
	$\nu_3$ is the grounded model, whereas $\nu_1$ and $\nu_2$ are both preferred and two-valued, and only $\nu_2$ is stable.
\end{example}

\section{Boolean Networks}\label{sec:bn}

In this section, we recall Boolean networks as known from the literature (see e.g., \cite{computingtrapspaces}), first looking at the syntax and then focussing on their semantics. 
During this presentation, we will establish sometimes surprising connections to ADFs as well as notable differences between these two formalisms.
\subsection{Syntax}
Boolean networks utilize a regulatory graph to represent the compounds in the biological process and the principal interactions between them. \begin{definition}
	A \emph{regulatory graph} is a directed graph \(G = (\){V}\(,E)\), where \(V = \{v_1,...,v_n\}\) is the set of vertices (nodes) representing the regulatory compounds, and \(E = \{(u,v,s): u,v \in V,  s \in \{+,-\}\}\) is the set of signed edges representing the \emph{interactions} between compounds. \end{definition}
An edge with $s=+$ is called \emph{positive interaction} (or \emph{activation}), representing that $u$ activates $v$, while an edge with $s=-$ is called \emph{negative interaction} (or \emph{inhibition}), representing that $u$ inhibits $v$.
A node with no incoming edges is called \emph{input node}, representing external stimuli, whose values do not change.

\begin{figure}[!t]
	\centering
	\ \ \ \ \ \	\includegraphics[scale=0.25]{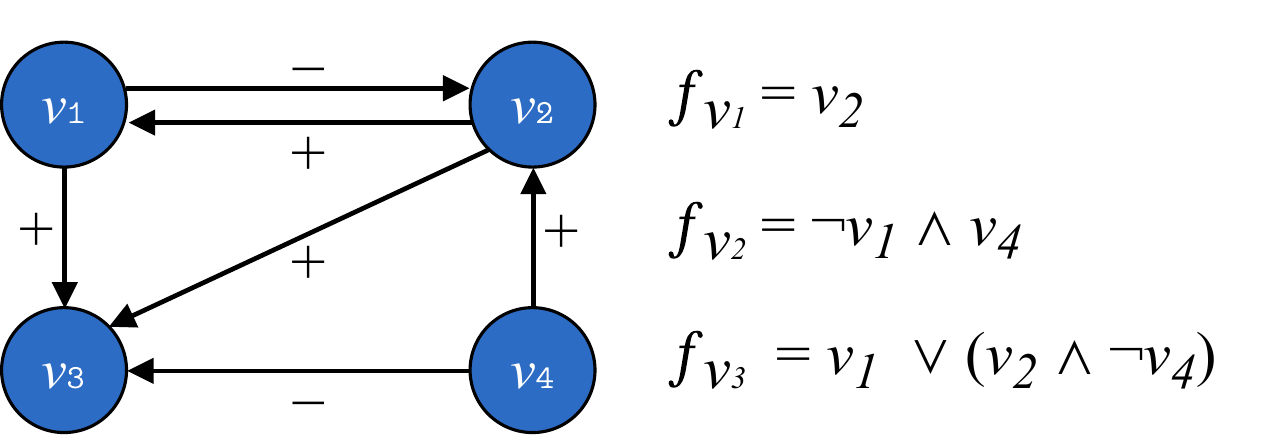}
	\caption{Example of a Boolean logical model.}
	\label{fig:completereggraphex}
\end{figure}

\begin{example}\label{ex:graph}
	Fig.~\ref{fig:completereggraphex} shows regulatory graph \(G = (\){V}\(,E)\) with \(V = \{v_1,v_2,v_3,v_4\}\) and \(E = \{(v_1,v_2, -),$ $(v_1,v_3, +), (v_2,v_1,+), (v_2,v_3, +), (v_4,v_2, +),(v_4,v_3, -)\}\).
\end{example}

Boolean logical models then add regulatory functions for each compound to specify how different compounds that affect the same node interact with each other for that node's activation.

\begin{definition}A \emph{Boolean logical model} \(M\) of a regulatory network is defined as a tuple \((\){V}\(,F)\) where \(V = \{v_1,v_2,...,v_n\}\) is the set of variables representing the regulatory compounds of the network such that \(v_i\) can be assigned to a value in $\{0,1\}$, and \(F=\{f_1, f_2,...,f_n\}\) is the set of Boolean functions such that \(f_i\) defines the value of \(v_i\) and where $f_i=v_i$ if $v_i$ is an input node. 
\end{definition}
Regulatory functions of input nodes may sometimes be omitted (cf.~Fig.~\ref{fig:completereggraphex}), which means that $f_i=v_i$, but commonly, we use the explicit representation.

\begin{example}\label{ex:model}
	Fig.~\ref{fig:completereggraphex} presents a boolean logical model with $G$ from Ex.~\ref{ex:graph} and regulatory functions for $G$ on the right.  
\end{example}

We can observe that, syntactically, a Boolean network is strikingly similar to an abstract dialectical framework. 
The only difference is that, unlike the edges in the regulatory graph of a BN, links in \ADFs do not mention explicitly whether an argument is attacking or supporting. 
Still, this can be extracted from the corresponding acceptance conditions.
We establish this connection formally.

\begin{definition}\label{def:conversion}
	Let $M=(V,F)$ be a Boolean logical model of a regulatory network with regulatory graph $G=(V,E)$. We define the corresponding \ADF $D_{M,G}=(V,L,F)$ with $L=\{(u,v)\mid (u,v,s)\in E\}$.
	
	Let $D=(\atoms,L,C)$ be an \ADF with $C$ in NNF. We define the corresponding Boolean logical model $M_D=(\atoms,C)$ of a regulatory network with regulatory graph $G_D=(\atoms,E)$ with $E=\{(u,v,+)\mid u\in C_v\}\cup \{(u,v,-)\mid \neg u\in C_v\}$.
\end{definition}
The requirement for the acceptance conditions to be in Negation Normal Form (NNF) is necessary to include the correct edges in $G_D$. Alternatively, one can determine the polarity using Definition \ref{def:polarity} below.

\subsection{Dynamics}

BNs allow us to capture the changes over time in a biological process based on the interactions of the various compounds involved, which should correctly represent the dynamics observable in the real system. We will see that the study of dynamics in BNs correspond to \emph{semantics} of \ADFs.
We start with network states that are used to represent the (current) activations of a network's compounds.

\begin{definition}
	The \emph{network state} of a BN with $n$ compounds is a vector \( S = [v_1,v_2,...v_n]\) where \(v_i\) is the value of the variable representing the $i$-th compound of the network.
\end{definition}
Clearly, for Boolean logical models, the number of different states in a network is given by \(2^n\).
E.g., if nodes $1$ and $3$ in Ex.~\ref{ex:model} are active and the other two are not, then the state will be represented by \(1010\). Similar representations are used for interpretations of \ADFs (by ordering the atoms), and it is clear that the states in BNs correspond to possible world of \ADFs.

The update of the i-th compound $v_i$ of a network from one discrete time point $t$ to the next is then defined as $v_i(t+1)=f_i(S(t))$ for state $S(t)$ at time $t$, which clearly exactly corresponds to the evaluation of an acceptance condition  in \ADFs.
We can then use state transition graphs  \cite{dynamicallyconsistentreductionoflogicalreggraphs} to describe how networks, and thus the modelled biological systems, evolve over time.

\begin{definition}
	A State Transition Graph (STG) is a directed graph \(G_{STG} = (S,T)\) where $S$ is the set of vertices representing the different states of the network, and $T$ is the set of edges representing the viable transitions between states.
\end{definition}

Two update schemes are employed to update the values of nodes in a BN: the synchronous and the asynchronous updating scheme \cite{dynanalysboolmammal,syncvsasyncgenregnets}.
Note that, given a Boolean logical model, the state transition graphs for the synchronous updating scheme and asynchronous updating scheme are uniquely determined.

\begin{figure}[!t]
	\centering
	\includegraphics[scale=0.15]{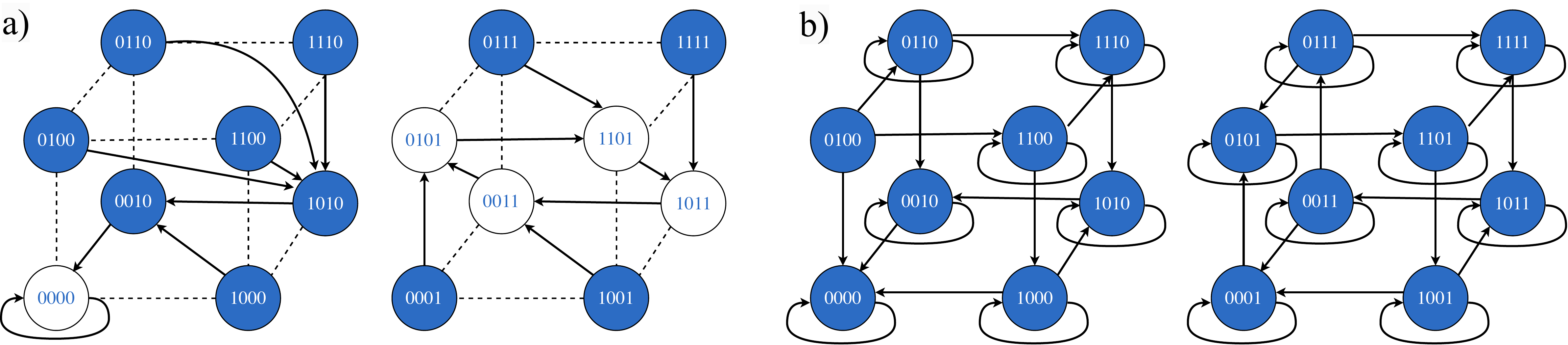}
\caption{{STG} of the model in Ex.~\ref{ex:model} for a) synchronous updates and b) asynchronous updates, both with \(v_4\) inactive on the left, and active on the right (adapted from \protect\cite{FilipeGouveia_9_2020}).
	}
	\label{fig:stg}
\end{figure}

In the synchronous updating scheme, at each time step, all compounds are updated simultaneously. Each network state has at most one successor (cf.~Fig.~\ref{fig:stg}.a), which is sometimes argued to be biologically less realistic and less accurate for analysis. Yet, synchronous updates are still regularly used \cite{SchwabEtAl20}, and many of the concepts below, such as trap spaces, are independent of the used scheme.

In the asynchronous updating scheme, at each time step, one or more regulatory functions may be applied \cite{SchwabEtAl20}. This is closer to what is observable in real systems, since these changes seldomly tend to take place simultaneously. With \emph{n} compounds in a network, and the frequently used, particular case of exactly one function being applied at each time step, each state can have at most \(n\) possible state transitions (including a transition to itself - cf.~Fig.~\ref{fig:stg}.b).

There are certain cycles of states in which these networks reside most of the time. These cycles are a trap of sorts, since as soon as a network enters a cycle's state, it is unable to exit the cycle. These traps, called \emph{attractors}, are linked to many important cellular processes, such as phenotypes, cell cycle phases, cell growth, differentiation, apoptosis, and more \cite{attractortutorial}. Because the number of states in a network is finite, transitions from all states eventually lead to an attractor. All states that lead to a certain attractor form its {attraction basin}.

Attractors have other relevant characteristics. One such characteristic is that, from any state belonging to an attractor, it is possible to find a path of attractor states to any other state in that attractor. Another important property is that there are no state transitions from attractor states to states outside of the attractor. Attractors can also be classified under different types. The left image of Fig.~\ref{fig:stg}.a) shows an example of a \emph{stable state} (or \emph{point attractor}) in the white-colored \emph{0000} state. Stable states are attractors that contain only a single state. When that is not the case, the attractor is denoted as a \emph{cyclic} (or \emph{complex}) \emph{attractor}, visualized in the right image of Fig.~\ref{fig:stg}.a). There, we have a complex attractor comprised of four states: \emph{0101}, \emph{1101}, \emph{1011} and \emph{0011} \cite{computingsteadystates}.

\begin{definition}
Given an STG \(G_{STG} = (S,T)\), a set $S'\subseteq S$ is a \emph{trap set of $G_{STG}$} if, for every $\omega\in S'$, $(\omega,\omega')\in T$ implies $\omega'\in S'$. An \emph{attractor} is a $\subseteq$-minimal trap set. A \emph{stable state of $G$} is a singleton trap set of $G$.
\end{definition}

We obtain that stable states of a transition graph $G$ under synchronous update scheme coincide with the two-valued models of the corresponding ADF.
\begin{propositionAprep}
Let $G_{STG}$ be the synchronous state transition graph of the Boolean Model $M$ with regulatory graph $G$, and $D_{M,G}$ the corresponding ADF.
	Then, $\omega$ is a stable state of $G_{STG}$ iff $\omega$ is a two-valued model of $D_{M,G}$.
\end{propositionAprep}
\begin{appendixproof}
	This follows from the following list of equivalences:
\begin{itemize}[nolistsep,noitemsep]
		\item[ ]$\omega$ is a stable state for the synchronous transition graph of $M=(V,F)$
\item[$\Leftrightarrow$] $s_i=f_i(\omega)$ for every $s_i\in\atoms$
		\item[$\Leftrightarrow$] $\omega(s)=\omega(C_s)$ for every $s\in\atoms$
		\item[$\Leftrightarrow$] $\omega$ is a two-valued model of $D$.
	\end{itemize}
\end{appendixproof}

To the best of our knowledge, the more general notion of a trap set has not been investigated in the context of \ADFs. This is not surprising, as it has a clear meaning and use in biological networks, but not so much in argumentation:
\begin{example}
	Consider the \ADF $D=(\{a,b,c\},L,C)$ with $C_a=\lnot c$, $C_b=\lnot a$ and $C_c=\lnot b$. Then $\{000,111\}$ is a trap set and an attractor, but their argumentative interpretation is not clear: if we interpret $a$, $b$, and $c$ as arguments that attack each other, the stability under transitions of  $\{000,111\}$, interesting in a biological interpretation, is of less interest in argumentation.
\end{example}

More surprisingly, we will now see that many other semantics for \ADFs have a natural counterpart in Boolean networks. For example, there has been a lot of interest in so-called \emph{subspaces} of regulatory graphs \cite{computingtrapspaces}, which are sets of interpretations for which assignments of some variables are fixed. 
Trap spaces have received a lot of attention in the literature on Boolean networks as finding them is computationally easier then finding \emph{any} trap set \cite{moon2022computational}, while they are still guaranteed to contain a trap set \cite{computingtrapspaces}.

\begin{definition}
	A subspace $m$ of a regulatory graph $(V,E)$ is a mapping $m:V\mapsto \{0,1,\star\}$. We call $v\in V$ \emph{fixed} if $m(v)\in \{0,1\}$ and \emph{free} if $m(v)=\star$, and $D_m$ is the set of all fixed variables in $m$. We let $S[m]:=\{ s\in S\mid \forall v\in D_m: s(v)=m(v)\}$.
\end{definition}

An interesting observation is that subspaces, interpreted as their representative set of boolean valuations $S[m]$, correspond exactly to the completions of the subspaces, interpreted as three-valued interpretations:
\begin{propositionAprep}\label{prop:spaces:are:completions}
	Let $m$ be a subspace $m$, and $\nu_m:V\mapsto \{0,1,{\sf U}\}$ defined by 
	\begin{eqnarray}
		\nu_m(v)=
		\begin{cases}
			m(v) & \mbox{ if }v\in D_m\\
			{\sf U } & \mbox{ otherwise}.
		\end{cases}
	\end{eqnarray}
	It holds that $[\nu_m]^2=S[m]$.
\end{propositionAprep}
\begin{appendixproof}
	It suffices to observe that $\nu_m\leq_i \possibleWorld$ iff $\possibleWorld(v)=m(v)$ for every $v\in D_m$.
\end{appendixproof}
Subspaces that are also trap sets are of special interest in the literature on Boolean networks.
\begin{definition}
	A \emph{trap space} is a subspace that is also a trap set.
\end{definition}

Trap spaces, interpreted as three-valued interpretations (see Proposition \ref{prop:spaces:are:completions}) correspond to admissible interpretations:
\begin{propositionAprep}\label{prop:admissible:iff:trapspace}
Let $M$ be a Boolean Model with regulatory graph $G$, and $D_{M,G}$ the corresponding ADF:
$m$ is a trap space  of M iff $\nu_m$ is admissible in $D_{M,G}$.
	
\end{propositionAprep}
\begin{appendixproof}
We first need some preliminaries due to Klarner et al.\ \cite{computingtrapspaces}. Given a boolean function $f$, 
	$f[m]$ is the expression obtained by stituting every occurence of some $v\in D_m$ in $f$ by $m(v)$. For example, given $m=0\star 1$ and $f=(a\lor b)\land c$, $f[m]=(0\lor 1)\land c$.
	We let $U_{m}:=\{v_i\in V\mid f_i[m]\mbox{ is constant}\}$ and define $F[m]:V\mapsto  \{0,1,{\sf U}\}$ as:
	\begin{eqnarray}
		F[m](v_i) =
		\begin{cases}
			f_i[m] & \mbox{ if }v_i\in U_{m}\\
			\star & \mbox{ otherwise }
		\end{cases}
	\end{eqnarray}
	The core of the proof depends on the following result:
	\begin{proposition}[\cite{computingtrapspaces}, Theorem 1]
		A space $m$ is a trap set if and only if $S[m]\subseteq S[F[m]]$.
	\end{proposition}
	Indeed, with Proposition \ref{prop:spaces:are:completions}, $[\nu_m]^2\subseteq [\nu_{F[m]}]^2$, i.e.\ $\nu_m\leq_i \nu_{F[m]}$. We now show that $\Gamma_{D_{M,G}}[\nu_m]=\nu_{F[m]}$.
	Recall that $\Gamma_{D_{M,G}}[\nu_m](v_i)=\sqcap_i\{ \possibleWorld(C_s)\mid \possibleWorld\in [\nu_m]^2\}$.
	We can safely assume that  $C_s$ is in conjunctive normal form, as  $\Gamma_{D_{M,G}}$ is invariant under classical equivalences. Let $C_s=\bigwedge_{i=1}^n \bigvee\Delta_i$. Then $f_s[m]=\bigwedge_{i=1}^n \bigvee\sigma(\Delta_i)$ where $\sigma(\Delta)$ is obtained by replacing every occurrence of some $v\in D_m$ by $m(v)$. Suppose now $F[m](s)=0$. This means there is some $i=1,\ldots,n$ s.t.\ $\sigma(\Delta_i)=\{0\}$. Thus, $\sqcap_i\{\possibleWorld(\bigvee\Delta_i)\mid \possibleWorld\in [\nu_m]^2\}=0$ which implies that $\sqcap_i\{\possibleWorld(\bigwedge_{i=1}^n \bigvee\Delta_i)\mid \possibleWorld\in [\nu_m]^2\}=0$. The cases for $F[m](s)=1$ and $F[m](s)=\star$ are similar.
\end{appendixproof}

On the other hand, the complete semantics does not seem to have a direct counterpart in Boolean networks.
\begin{example}
	Consider the \ADF $D=(\{a,b\},L,C)$ with $C_a=\{a\lor \lnot a\}$ and $C_b=b$. Let us take a look at the STG for synchronous updates first:
\begin{center}
\includegraphics[scale=0.25]{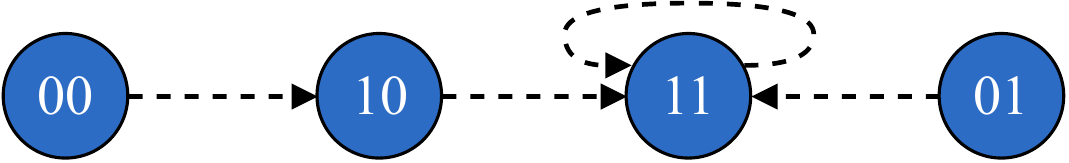}
\end{center}
There are three trap spaces: $\star\star$, $\star1$, and $11$. By Prop.~\ref{prop:admissible:iff:trapspace}, this corresponds to the interpretations $\undec\undec$ and $11$. However, the unique complete interpretation is $11$. 
\end{example}
We might conjecture that the complete interpretations correspond to the minimal trap spaces, but that is not the case either.

\begin{example}
	Consider the \ADF $D=(\{a\},L,C)$ with $C_a=\{a\}$. 
	There are three trap spaces: $\star$, $1$, and $0$. The corresponding interpretations $\undec$, $1$ and $0$ are not only admissible but also complete. However, $\undec$ is not a minimal trap space.
\end{example}

We further note that there has been some interest in maximal and minimal trap spaces \cite{computingsteadystates} in the literature on \BNs. 
In more detail, trap sets are compared as follows: $m_1\leq m_2$ iff $S[m_1]\subseteq S[m_2]$.
 It can be easily observed that this is the reverse of the information order $\leq_i$ known from \ADFs:
\begin{propositionAprep}\label{prop:information:order}
Let $m_1,m_2$ be two subspaces. Then $m_1\leq m_2$ iff $\nu_{m_2}\leq_i \nu_{m_1}$.\nocite{heyninck2020epistemic}
\end{propositionAprep}
\begin{appendixproof}
With Proposition 2 from \cite{heyninck2020epistemic},
$\nu_{m_2}\leq_i \nu_{m_1}$ iff for every $s\in\atoms$ iff $[\nu_{m_2}]^2\supseteq [\nu_{m_1}]^2$.
As $S[m_i]=[\nu_{m_i}]^2$ for $i=1,2$ (Proposition \ref{prop:spaces:are:completions}), we obtain that 
$\nu_{m_2}\leq_i \nu_{m_1}$ iff $S[m_1]\subseteq S[m_2]$. By definition of $\leq$, $\nu_{m_2}\leq_i \nu_{m_1}$ iff $m_1\leq m_2$.
\end{appendixproof}
\emph{Minimal} trap spaces are trap spaces $m_1$ such that there is no trap space $m_2<m_1$. \emph{Maximal} trap spaces are trap spaces $m_1$ s.t.\ there is no trap space $m_2>m_1$ and $S[m_1]\neq S$, i.e., the trivial trap space is excluded by definition. 
\begin{propositionAprep}
Let $M$ be a \BN with regulatory graph $G$, and $D_{M,G}$ the corresponding ADF:
$m$ is a minimal trap space of M iff $\nu_m$ is preferred in $D_{M,G}$.
\end{propositionAprep}
\begin{appendixproof}
With Proposition \ref{prop:admissible:iff:trapspace}, $m$ is a trap space of M iff $\nu_m$ is admissible in $D_{M,G}$. Suppose now towars a contradiction that $m$ is a minimal trap space, yet $\nu_m$ is not preferred, i.e.\ there is some admissible $\nu$ with $\nu_m<_i \nu$. Then with Proposition \ref{prop:information:order} and proposition  \ref{prop:admissible:iff:trapspace}, there is a trap space $m_\nu< m$, contradicting $m$ being a minimal trap space.
\end{appendixproof}
However, maximal trap spaces do not correspond to the grounded model. The reason is that in \BNs, the trivial trap space is excluded. E.g., in Ex.~\ref{exa:ADF:1}, $\nu_3$ is the grounded model, but does not correspond to a maximal trap space as it is trivial trap space.
Finally, we note that the concept of stable model from \ADFs has no clear counterpart in \BNs. Indeed, the main motivation behind stable models is to exclude self-supporting arguments (see e.g., Ex.~\ref{ex:travelMods} where $\nu_1$ with $p$ supporting itself via $C_p=p$ is excluded). In \BNs, there is nothing \emph{a priori} wrong with such self-supporting stable states, and it might often even have a clear biological meaning, e.g., since algae are self-reproducing. 

\subsection{Subclasses of Boolean Networks}\label{sec::classes}
An interesting observation is that both the literature on \ADFs and the literature on Boolean Networks has identified certain subclasses of frameworks for which the computational complexity of computational tasks decreases. In fact, both strands of literature have identified the same subclass! In the literature on \ADFs, these frameworks are called \emph{bipolar} ADFs, whereas in the literature on Boolean networks, they are called \emph{sign-definite}.

\begin{definition}
	A Boolean function $f: \{0,1\}^n\mapsto \{0,1\}$ is \emph{increasing monotone} if for every $v_1,v_2\in \{0,1\}^n$, $v_1\leq v_2$ implies $f(v_1)\leq f(v_2)$, and it is  \emph{decreasing monotone} if for every $v_1,v_2\in \{0,1\}^n$, $v_1\leq v_2$ implies $f(v_1)\geq f(v_2)$. A Boolean function is \emph{sign-definite} if it is increasing or decreasing monotone. A Boolean logical model $(V,F)$ is \emph{sign-definite} if every Boolean function in $F$ is sign-definite.
\end{definition}
It is well-known that a Boolean function is sign-definite if and only if it can be represented by a formula in disjunctive normal form in which all occurrences of a given literal are either negated or non-negated \cite{anthony2001discrete}.

We now recall \emph{bipolar} \ADFs \cite{strass2013approximating}.
\begin{definition}\label{def:polarity}
	Given a Boolean function $f:\{0,1\}^n\mapsto \{0,1\}$:
	\begin{itemize}
		\item $i\leq n$ is \emph{supporting} iff $f(v_1,\ldots, v_i,\ldots,v_n)=1$ implies  $f(v_1,\ldots, v'_i,\ldots,v_n)=1$ where $v_i=0$ and $v'_i=1$,
		\item $i\leq n$ is \emph{attacking} iff $f(v_1,\ldots, v_i,\ldots,v_n)=0$ implies  $f(v_1,\ldots, v'_i,\ldots,v_n)=0$ where $v_i=0$ and $v'_i=1$.
	\end{itemize}
	An \ADF is \emph{bipolar} iff for every $a\in \atoms$, every $b\in \mathrm{par}_D(a)$ is supporting, attacking or both in $C_a$.
\end{definition}

It was shown that an \ADF is bipolar iff every acceptance formula is syntactically bipolar, i.e., no atoms occurs both positively and negatively \cite{strass2015expressiveness}. 
In more detail, given a formula $\phi$, the \emph{polarity} of an atom $a$ in $\phi$ is determined by the number of negations on the path from the root of the formula tree to the atom, and is positive if this number is even, and negative otherwise. For example, in $\lnot (a\land \lnot b)$, $a$ occurs negatively and $b$ occurs positively. 
A propositional formula is \emph{syntactically bipolar} if no atom $a$ occurs both positively and negatively in $\phi$. 
\begin{corollaryAprep}
An \ADF $D=(\atoms,L,C)$ is bipolar iff, for every $s\in\atoms$, $C_s$ is syntactically bipolar.
\end{corollaryAprep}
\begin{appendixproof}
Follows from \cite[Theorem 1]{strass2015expressiveness}.
\end{appendixproof}
\begin{toappendix}
The reader might be somewhat surprised by the fact that links can be both attacking and supporting. However, the only possibility for that being the case, is if the link is redundant:
\begin{propositionAprep}
Consider an \ADF $D=(\atoms,L,C)$ with $s_1,s_2\in \atoms$ and $s_2$ being supporting and attacking in $C_{s_1}$.
For every $v_1,v_2\in \mathcal{V}^2(\atoms)$ s.t.\ $v_1(s_2)\neq v_2(s_2)$ and $v_1(s)=v_2(s)$ for every $s\in \atoms$, $\Gamma_D(v_1)(s_1)= \Gamma_D(v_1)(s_1)$.
\end{propositionAprep}
\begin{appendixproof}
Wlog supose that $v_1(s_2)=0$ and $v_2(s_2)=1$.
Suppose first that $C_{s_1}(v_1)=1$. As $s_2$ is supporting in $C_{s_1}$,  $C_{s_1}(v_2)=1$.
Suppose now that $C_{s_1}(v_1)=0$. As $s_2$ is attacking in $C_{s_1}$,  $C_{s_1}(v_2)=0$.
\end{appendixproof}
\end{toappendix}
This allows us to show that these two special cases coincide.
\begin{corollaryAprep}
A Boolean Model is sign-definite iff   the corresponding ADF is bipolar.
\end{corollaryAprep}
\begin{appendixproof}
Clearly, an acceptance condition is supporting respectively attacking iff it is increasing respectively decreasing monotone. 
\end{appendixproof}

\section{Insights from Boolean Networks}\label{section:results:from:bns}
Based on the established correspondence, we provide examples of insights from the literature on Boolean Networks that are directly relevant for \ADFs.

\subsection{Complexity of Counting}
For many applications in KR, counting the number of solutions is important. For many formalisms, including abstract argumentation \cite{fichte2019counting}, the complexity of counting has been studied in depth. For \ADFs, such as study is missing. Thanks to existing results from the literature on \BNs we can start filling in this gap:
\begin{proposition}[\cite{bridoux2022complexity}]
	Deciding whether the maximal number of stable states in a Boolean logical model $M$ is larger or equal than $k$ is:
 {\sf NP}-complete for $k\geq 2$;
{\sf NEXPTIME}-complete for an arbitrary but fixed $k$;
{\sf NP}$^{\#{\sf P}}$-complete for an arbitrary but fixed $k$ if the maximum in-degree of $M$ is bounded by a constant $d\geq 2$.

	Deciding whether the minimal number of stable states in a Boolean logical model $M$ is smaller or equal than $k$ is:
{\sf NEXPTIME}-complete for an arbitrary but fixed $k$;
{\sf NP}$^{\sf NP}$-complete for $k\geq 2$, if the maximum in-degree of $M$ is bounded by a constant $d\geq 2$;
{\sf NP}$^{\#{\sf P}}$-complete for an arbitrary but fixed $k$ if the maximum in-degree of $M$ is bounded by a constant $d\geq 2$.
\end{proposition}

This gives rise to the following corollary:
\begin{corollary}
	Deciding whether the maximal number of two-valued models in an \ADF $D$ is larger or equal than $k$ is:
  {\sf NP}-complete for $k\geq 2$;
 {\sf NEXPTIME}-complete for an arbitrary but fixed $k$;
{\sf NP}$^{\#{\sf P}}$-complete for an arbitrary but fixed $k$ if the maximum in-degree of $D$ is bounded by a constant $d\geq 2$.
	
	Deciding whether the minimal number of two-valued models in an \ADF $D$ is smaller or equal than $k$ is:
 {\sf NEXPTIME}-complete for an arbitrary but fixed $k$;
 {\sf NP}$^{\sf NP}$-complete for $k\geq 2$, if the maximum in-degree of $M$ is bounded by a constant $d\geq 2$;
 {\sf NP}$^{\#{\sf P}}$-complete for an arbitrary but fixed $k$ if the maximum in-degree of $D$ is bounded by a constant $d\geq 2$.
\end{corollary}

\subsection{Existence of Fixpoints}
There is a large line of work in the literature on Boolean networks that studies structural properties of Boolean networks that affect the existence (and the number) of fixed points. Such work immediately translates to the existence of two-valued models in \ADFs. E.g., several works \cite{robert1980iterations,remy2008graphic,richard2010negative,aracena2008maximum}
provide a theoretical analysis of the existence of fixpoints in sign-definite BN in relation to structural parameters of the corresponding Boolean networks.
We follow Aracena \cite{aracena2008maximum} in defining a path in a BN as \emph{positive} if the number of negative arcs is even, and negative otherwise.

The first few results study the connection between cycles, their parity, and the existence and number of fixpoints:

\begin{proposition}If a sign-definite BN has:
no cycles, then it has a unique stable state \cite{robert1980iterations};
no positive cycles, then it has at most one stable state \cite{remy2008graphic};
no negative cycles, then it has at least one stable state \cite{richard2010negative}.
 If a BN $N$ has a strongly connected component $H$ such that all cycles of $H$ are negative and for each arc $(v_i,v_j )$ in the $N$, $v_j\in H$ then $v_i\in H$, then $N$ has no stable states \cite{aracena2008maximum}.
\end{proposition}

We derive the following corollary for \ADFs:
\begin{corollary}
	Let a bipolar \ADF $D$ be given. Then:
 if $D$ has no cycles, then it has a unique two-valued model;
 if $D$ has no positive cycles, then it has at most one two-valued model;
 if $D$ has no negative cycles, then it has at least one two-valued model.
If $D$ has a strongly connected component $H$ such that all cycles of $H$ are negative and for each arc $(v_i,v_j )$ in the $N$, $v_j\in H$ then $v_i\in H$, then $N$ has no two-valued models.
\end{corollary}
One finds also more intricate results on the existence of fixpoints in the literature, for example in terms of so-called non-expansive maps, that guarantee the existence of stable states \cite{richard2011local}. These fall outside the scope of this paper.

Some work has also investigated the connection between the existence of stable states and the structure of the corresponding BN:
\begin{proposition}[\cite{aracena2008maximum}]
	If a BN has at least one stable state, then it has at least one positive cycle.
\end{proposition}
We derive the following corollary for \ADFs:

\begin{corollary}
	If an \ADF has a two-valued model, then it has at least one positive cycle. 
\end{corollary}

The final result we discuss is an upper bound on the number of stable states in terms of the number of \emph{feedback vertex sets} (FVSs). An FVS of a digraph $G = (V,E)$ is defined to be a set of vertices that contains at least one vertex of each cycle of $G$. The minimum number of vertices of a FVS is denoted by $\tau(G)$.
\begin{proposition}[\cite{aracena2008maximum}]
	A Boolean logical model $N=(V,F)$, where $|V^{-}(v)| \geq 1$ for all $v\in V$, has at most $2^{\tau(N)}$ stable states.
\end{proposition}

This carries over to \ADFs as follows.

\begin{corollary}
	An \ADF $D=(\atoms,L,C)$ s.t.\ every argument has at least one attacker has  at most $2^{\tau((\atoms,L))}$ stable states.
\end{corollary}
 	
	\section{Conclusion}\label{sec:conclusion}
We have reviewed the main syntactic similarities between \ADFs and \BNs, and demonstrated that these extend in large parts to the semantical level. Furthermore, we have shown the fruitfulness of this connection by deriving results on complexity and existence of semantics for \ADFs based on existing results for \BNs. We hope that our results will lead to further cross-contamination between these two fields, such as the use of implementations or benchmarks.
\bibliographystyle{plain}
\bibliography{bibliography}

\end{document}